# Data Mining for Prediction of Human Performance Capability in the Software-Industry

By: Gaurav Singh Thakur[1], Anubhav Gupta[2] and Sangita Gupta[3]

[1]Cisco Systems, Bangalore email: sai007gaurav@gmail.com,

[2]Common Floor Technologies, Bangalore email: anubhav992@gmail.com

[3]Jain University, Bangalore email: sangita101166@gmail.com

## Abstract

*The recruitment of new personnel is one of the most essential business processes which affect the quality of human capital within any company. It is highly essential for the companies to ensure the recruitment of right talent to maintain a competitive edge over the others in the market. However IT companies often face a problem while recruiting new people for their ongoing projects due to lack of a proper framework that defines a criteria for the selection process. In this paper we aim to develop a framework that would allow any project manager to take the right decision for selecting new talent by correlating performance parameters with the other domain-specific attributes of the candidates. Also, another important motivation behind this project is to check the validity of the selection procedure often followed by various big companies in both public and private sectors which focus only on academic scores, GPA/grades of students from colleges and other academic backgrounds. We test if such a decision will produce optimal results in the industry or is there a need for change that offers a more holistic approach to recruitment of new talent in the software companies. The scope of this work extends beyond the IT domain and a similar procedure can be adopted to develop a recruitment framework in other fields as well. Data-mining techniques provide useful information from the historical projects depending on which the hiring-manager can make decisions for recruiting high-quality workforce. This study aims to bridge this hiatus by developing a data-mining framework based on an ensemble-learning technique to refocus on the criteria for personnel selection. The results from this research clearly demonstrated that there is a need to refocus on the selection-criteria for quality objectives.*

## I. Introduction

Software engineering is a discipline of science which deals with systematic development of software. It has various methods, process and tools to achieve high quality products. However software-quality still isn't up-to the mark. Practitioners are realizing that people are the biggest risk if not performing well. Software quality can be achieved by cumulative quality standards of process and people over all generic modules in development [10]. It is utmost important to look into the people component deeply in the people, process, product triad. Data mining is the process of extracting useful knowledge from data [8]. It utilizes a combination of a knowledge base, sophisticated analytical skills and domain specific knowledge to uncover many hidden trends and patterns. These patterns and relations can be extracted by using various data-mining algorithms. Today, the human aspects of software engineering have become one of the critical concerns in IT companies to achieve their business goals. Industries be it in any sector, are now paying attention to selecting the right talent who can perform consistently well throughout all generic framework activities and execute the process properly. Software quality is greatly dependent on the people and process quality during the development and testing phase. Hence, in this paper we use data mining algorithms to exploit the patterns in the historical data and predict the performance based on project-personnel attributes and thereby enhance the quality of process and assure project success. Data Mining is the next big revolutionary field that is redefining the industry, be it in terms of technology or research. Here we use mathematical procedures like function-approximation techniques to solve prediction or estimation problems and extract useful trends and patterns from past data to facilitate ourselves to take right decisions with the aim to produce near-optimal results. Classification, allows us to identify association rules. Categorization uses induction algorithm rules to handle categorical outcomes, such as good, average and poor as in this study.

In this paper we address the issue of developing an ideal selection framework for recruiting the right talent which brings us to the basic question of what criteria to follow for the selection procedure. Our aim is to understand the relationship between the various project-personal attributes of the candidates and their professional-performance parameter as rated by their managers/supervisors in the industry. Our aim is to identify the variables which have the maximum predictive power in estimating the performance capabilities of new recruits. Though there have been many previous studies in this domain, there have been certain issues that still need to be addressed. We try to build upon them and use this research to build a better and more robust model that can be not be applied to different scenarios but also work well on different data sets having varied properties with minimal or no changes. There are a wide range of algorithms in each of these categories, many of which are implemented on WEKA and R [7]. These tools are platforms with GUI and command-line implementations respectively, with a number of machine-learning algorithms for data mining tasks, with a variety of options for regression, classification, data pre-processing, association rules, clustering and visualization.

The paper is organized as follows: Section 2 provides the related work and background of data mining Algorithms. Section 3 presents the research methodology to derive at a conclusion. Section 4 discusses about the implementation details while Section 5 depicts the obtained result. Section 6 discusses the summary of this paper and its future scope.

## II. Related Work

With increasing complexity of software in the industry and their ever growing demands in multidisciplinary projects, there has been a continuous progress in research works that target the areas of effective project management and Data mining has recently proven itself as one of the most established techniques in this area. Data mining methodologies are developed for several applications including various aspects of software development and we plan to employ this power of algorithms to develop a selection framework. There have been a plethora of studies which incorporate the tools of machine-learning for developing a framework for Prediction of Human-Capability.

The authors in [13] have studied the importance of different variables that come into play during the selection of students - like academic scores , Programming Skills, Domain Knowledge Assessment, Reasoning skills, Mental Ability and Mathematical skills and many more using Decision Trees ID3, CART and C4.5. As it turns out, there are many features that must be tested along with academic scores to confirm the quality of new personnel. However, the model used in this project has significant scope for improvement and we base our project on the hypothesis similar to this research by Sangita Gupta et. al. to further improve upon the results obtained using a different approach that involves Ensemble-learning technique – Random Forest, a bagging based approach to accuracy boosting. Focusing on other works, authors in [4] surveyed different machine learning algorithms for defect prediction in software. Authors of [5] have used data mining technique clustering and done a comparative analysis of performance of Density-Based Spatial Clustering of Applications with Noise (DBSCAN) and Neuro-Fuzzy System for predicting the level of severity of faults in Java-based object oriented software. Data mining results in decision through methods & not through assumptions. Authors in [2] have worked on the improvement of employee selection in semiconductor industry by developing a model, using data mining techniques. The specified attributes involved age, gender, marital-status, experience, education, major-subjects and school-tires as potential factors that might affect the performance. As an outcome of their study, it was found that employee-performance is greatly affected by educational-degree, the school-tire, and the work experience. The authors in [3] researched on multiple factors that affect the job performance of employees. They reviewed previous works which study the effect of experience, salary and training, working-conditions and job satisfaction on the performance-parameters. Data-mining thus supports various techniques including Statistical-Analysis, Decision-Trees, Genetic-Algorithm, Bayes-classification, visualization techniques, etc. for analysis and prediction. It further facilitates association, clustering and classification [1]. This research-study involves applying data in WEKA and R tool and derives a classification model for selection criteria. The algorithm we use in this project is Random-Forest, a bagging based accuracy boosting technique which improves upon the result published in [13] which uses Decision Trees namely ID3, C4.5 and CART.

## III. Research Methodology

The research methodology followed during this study involved the following steps as stated below:

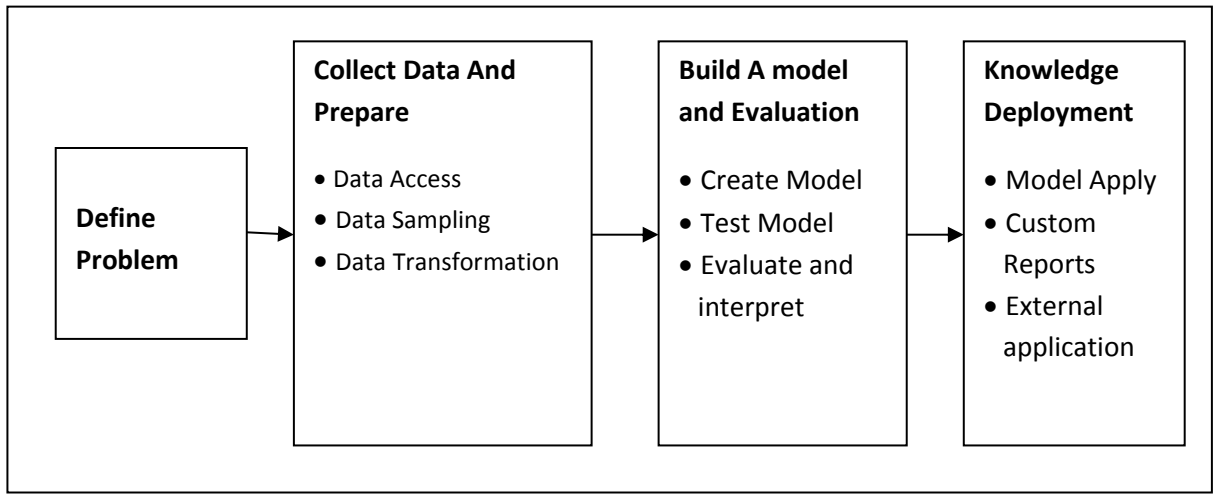

**Figure 1 : Framework for Building recruitment Process**

- **Hypothesis**: Project personnel with similar skills set and capabilities will perform similarly.
  Project-personal information about the employees is collected from projects which use similar technology and programming language and work on similar platforms. Thus personnel under comparison here have similar capabilities and skill-sets.

- **Data Collection**: Data was collected by using various techniques such as Form-Filling, Brainstorming, obtaining performance information about employees from the Project-leads and managers.

- **Data Preparation**: A basic preliminary research concluded with an almost same set of attributes that must be considered under this problem-statement to obtain a correlation with performance parameters as in [13]. The list of attributes noted are as follows:

  I. **PS** - Programming Skills: The programming and coding skills of employees were tested using standing coding problems with multiple test cases. A complete score was awarded for each problem only if all the test cases were executed successfully. Employees were tagged as good for scoring above an overall value above 75%, average if they scored between 50% and 75% and the rest were tagged as poor.

II. **RAS –** Reasoning and Analytical Skills: Similar to programming skills, RAS values were collected through an internal assessment which involved analytical and logical-reasoning questions.

III. **DSK** - Domain Specific Knowledge: Categorized and normalized similar to PS and RAS. Obtained from a series of internal assessments, DSK refers to both theoretical and field-based knowledge about the domain in which the employees have been working.

IV. **TE** - Time efficiency**:** Marked as a simple "Good" or "Bad" option by the project manager/leader of an employee which shows if the employee is considered efficient in completing his/her work within the expected time-period.

V. **GPA** – Grade Point Average: This is the grade scored by an employee during his/her graduation/ post-graduation courses. It is categorized and normalized to scale the GPAs on the same level based on universities.

VI. **CS** - Communication Skills: This score for each of the employees was obtained from the project-leaders, their team mates and the respective HR-manager of the employee as well.

VII. **P** – Performance Parameter: This is our dependent variable for the problem statement which must be correlated to the independent attributes available from the preliminary study of relevant features. The value was acquired by brainstorming with the Team-leads and Managers assigning an overall performance score to the employees in terms of good, average and poor.

This preliminary-study of relevant attributes is followed by the selection of a suitable machine-learning model that must be adopted for the data-mining process. As we have seen, a number of studies have already been done in this domain but the issue has been lack of high-accuracy. Such accuracy issues can be addressed very well if accuracy-boosting techniques are applied. Also given the fact that Knowledge-based Decision Trees have been known to perform decently in such cases, we employ the bagging based Ensemble-learning model Random forest in this scenario to further enhance the performance of old models.

**Underlying Learning Model - Random Forest**

In a classification problem, we have a training sample of n observations on a class variable Y that takes values 1, 2, ... k, and p predictor variables, X1,..., Xp. Our goal is to find a model for predicting the values of Y from new X values. In theory, the solution is simply a partition of the X space into k disjoint sets, A1, A2,..., Ak, such that the predicted value of Y is j if X belongs to Aj , for j = 1, 2,..., k. If the X variables take ordered values, two classical solutions are linear discriminant analysis1 and nearest neighbor classification. These methods yield sets Aj with piecewise linear and nonlinear, respectively, boundaries that are not easy to interpret if p is large. Classification tree methods yield rectangular sets Aj by recursively partitioning the data set one X variable at a time. This makes the sets easier to interpret. For example, Figure 1 gives an example wherein there are three classes and two X variables. The left panel plots the data points and partitions and the right panel shows the corresponding decision tree structure. A key advantage of the tree structure is its applicability to any number of variables, whereas the plot on its left is limited to at most two.

Pseudocode for tree construction by exhaustive search is as follows:

- Start at the root node.

- For each X, find the set S that minimizes the sum of the node impurities in the two child nodes and choose the split $\{X* \in S*\}$ that gives the minimum overall X and S.

- If a stopping criterion is reached, exit. Otherwise, apply step 2 to each child node in turn.

Since, this is a Regression problem, we use CART for producing individual regression trees and then build upon this model by using the Random Forest technique to achieve better results. CART uses a generalization of the binomial variance called the Gini index. A regression tree is similar to a classification tree, except that the Y variable takes ordered values and a regression model is fitted to each node to give the predicted values of Y. This yields piecewise constant models. Although they are simple to interpret, the prediction accuracy of these models often lags behind that of models with more smoothness. It can be computationally impracticable, however, to extend this approach to piecewise linear models, because two linear models (one for each child node) must be fitted for every candidate split.

These trees may not provide very high accuracies, since they have very high variance values. Randomization based ensemble methods, prove to be a good solution to this flaw. Random-Forest consists of a collection or ensemble of simple tree predictors, each of which outputs a response when presented with a set of predictor values just as the input vector X. For classification-based problems, this response can be of the forms - class membership or associations, a set of independent predictor values with one of the categories present in the dependent variable. Each tree is created from its own separate bootstrapped sample training set. The *Bootstrap Sampling Method* samples the given training tuples uniformly with replacement i.e. each time a tuple is selected, it is equally likely to be selected again and rendered to the training set. As the number of simple learning models within an ensemble technique increases, the overall variance of the output-value from the actual-value theoretically decreases by 1/(number of individual models). However this decrease in variance after a threshold doesn't yield significant improvement and that allows us to decide the number trees we want to create for this random forest technique. It's important to remember that ensemble learning techniques are computationally expensive and hence choosing an optimal value for the number of individual simple predictors within the ensemble technique is a critical task. Individual Decision-trees usually suffer from high-variance, which makes them uncompetitive in terms of accuracy. A highly efficient and simple way to address this issue is to adopt the context of randomization and use them in ensemble-methods.

The Mean-Decrease-in-Accuracy of a variable is evaluated during the calculation-phase of out-of-bag error. As the fall in accuracy of the random-forest increases due to the addition of a single-variable, the more important the particular variable under test is considered and hence variables with a large value for Mean-Decrease-in-Accuracy or Gini are considered as more important for data classification. The Mean-Decrease-in-Gini coefficient is a measure of how each particular variable supplements to the homogeneity of the nodes and terminal-leaves in the resulting Random-forest. Every time one particular variable splits a given node, the Gini coefficient for the children are calculated and compared to that of the original parent node. If the same variable causes multiple splits more than once, then the final difference in the Gini value of the topmost parent node and the bottom-most children nodes is taken as the Mean-decrease-in-Gini value. The pseudo code for generation of a random forest is as follows:

i.  Draw *number_of_trees* bootstrap samples from the original data.

ii. Grow a tree for each bootstrap data set. At each node of the tree, randomly select *mtry* variables for splitting. Grow the tree so that each terminal node has no fewer than *nodesize* cases.

iii. Aggregate information from the *ntree* trees for new data prediction such as majority voting for classification.

iv. Compute an out-of-bag (OOB) error rate by using the data not in the bootstrap sample.

Using these values, a final graph for *Variable Importance* is plotted, where this graph represents each variable on the vertical y-axis, and their importance-values on the horizontal x-axis. They are ordered in the manner of top-to-bottom as maximum-to-minimum-importance. To measure the accuracy of the classifier, we made use of Sensitivity and Specificity parameters. The following are the meaning of the variables used in the subsequent equations.

- True positive = correctly identified
- False positive = incorrectly identified
- True negative = correctly rejected
- False negative = incorrectly rejected

1. **True-Positive Rate** or Sensitivity is the fraction of training samples predicted correctly by model.
$$\text{TPR} = \frac{TP}{TP+FN} \quad (1)$$

where TPR represents the True-positive-Rate and higher this value, the better the model is.

2. **False-Positive Rate** or Specificity i.e. the fraction of training samples predicted incorrectly by model.
$$\text{FPR} = \frac{FP}{TN+FP} \quad (2)$$

where FPR represents the False positive Rate and lower this value, the better the model is.

3. **Area under ROC curve (Receiver Operating Characteristics)**: is obtained by plotting TPR against FPR. The area under the plotted graph gives a good measure of the accuracy of the classifier. The area can be as high as 1 Sq. unit (maximum accuracy) and as low as 0 (minimum accuracy). Since Area-Under-ROC-curve takes into consideration, both FPR and TPR values, this measure is preferred over the parameters to compare accuracies between models.

## IV. Implementation Details

Based on the study conducted, data obtained was consolidated and summarized in a tabular form. The algorithm used here is Random Forest, implemented using WEKA and R under "Test options". 10 - Fold Cross validation was applied to supplement the out-of-bag calibration mechanism of Random-Forests. The number of individual predictor-trees was set to 500 in R as no significant reduction in variance was observed beyond this value. Trees were allowed to grow completely without any pre or post pruning. The package used in R for implementing Random Forest is the "randomForest" package which is compatible with versions 4.6 and above. This package is available for download at the official R support website.

For creating bootstrap samples, we used the technique of 632 Bootstrapping, which means that in any bootstrap sample generated, approximately 63.2% of the Dataset will be unique, and the rest would be placed with replacement and duplication. Studies have shown that this bootstrapping technique produces near-optimal results. The plot for Variable Importance has been obtained using R tool and the results obtained are discussed below. The model generated cannot be visualized graphically due to the large number of trees generated, each created from a separate bootstrap sample and each producing its own results. As stated earlier, the final output from the Random Forest model is the average of the results obtained from each of the predictor trees in the forest.

# V. Results and Discussion

Table 1 shows the results obtained by Random forest. Table 2 shows the Variable Importance values for each of the attributes in terms of Mean-Decrease in Gini and Mean-Decrease in Accuracy.

| CLASS | TP RATE | FP RATE | Area under ROC curve |
|---|---|---|---|
| Good | 0.934 | 0.037 | 0.977 |
| Average | 0.846 | 0.037 | 0.983 |
| Poor | 0.929 | 0.077 | 0.992 |

Table 1 : Accuracy Measures for samples classified by output class

Firstly we look at the average area under the ROC curve and as we can see, this area is about 0.984 (average). The area here is close to 1 and for ROC curves; an area 1 refers to the highest possible accuracy of 100%. Hence, we can see that the Random-Forest Model used is highly accurate and strong conclusions can be drawn from these results. We also found through a comparative study that the model outperformed knowledge-based decision trees and Linear Regression techniques when applied to the same data.

We now take a look at the Variable importance plot using Mean-Decrease in Gini and Mean-Decrease in Accuracy values.

|  | DSK | RAS | PS | CS | TE | GPA |
|---|---|---|---|---|---|---|
| Mean Decrease Accuracy | 35 | 10 | 9 | 0 | 0 | 0 |
| Mean Decrease Gini | 14 | 5 | 4 | 2 | 1 | 2 |

Table 2 : Variable Importance measures for each Attribute

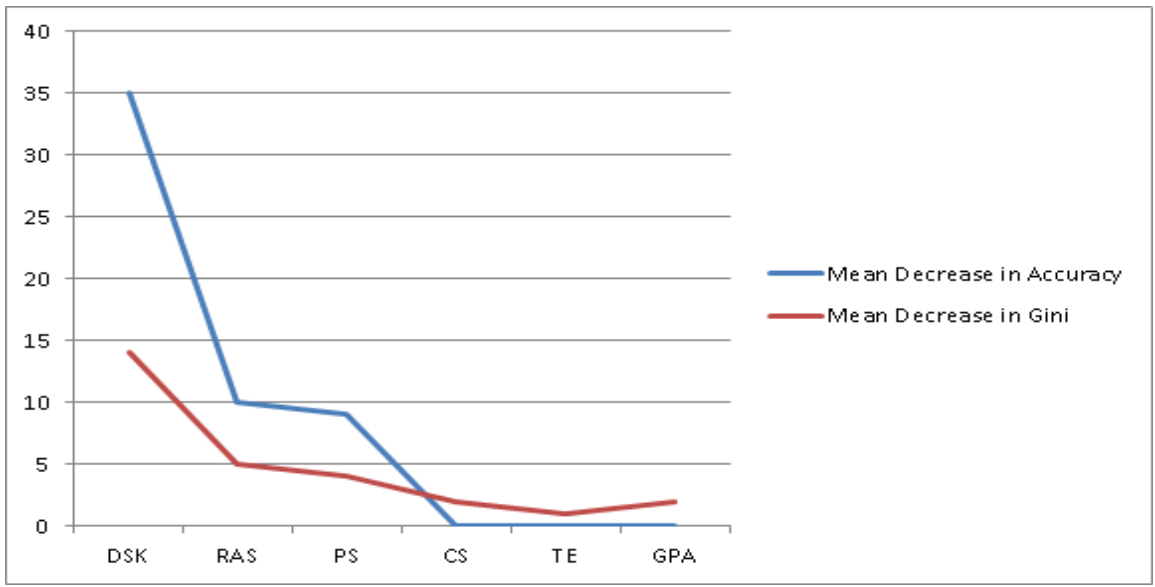

Figure 2 Plot for Variable Importance of different attributes

As we see in these plots, the variables like Domain-Specific knowledge, Reasoning and Analytical skills and Programming Skills are the most important attributes to be considered during recruitment of new personnel, as they have the maximum contribution towards the homogeneity of nodes and classification of data. Another important result obtained in this scenario is the Mean Decrease in Accuracy value for GPA which comes out to be 0. Such low scores for these values stand as a scientific base to challenge the usual recruitment procedure where maximum importance is given to GPA of candidates. What we see here is that, based on the grades and academic scores of students from universities, one cannot predict their performance in the industry. Hence GPA alone is not a very clear reflection of the candidate's capabilities as far as the software industry is concerned. It is needed that the companies separately test the other relevant attributes of the students in order to make a better decision. The software development process involves various intricate steps and complex stages where many other factors and abilities of an individual come into play. Referring to the GPA alone will not yield optimal results and this is the reason, why there has been a changing trend in the recruitment procedure today. Recruiting teams are looking for candidates with a complete package in terms of overall personality, analytical thinking abilities and good inter-personal skills apart from good grades. The importance given to personal interviews lays emphasis on the fact that only after getting a true idea about the candidate's knowledge in the particular domain through a one-on-one interaction, companies take a final decision regarding the selection procedure.

Hence, using the results from the variable importance plot, we create ideal knowledge based procedure consisting of rules which can be used as the selection criteria in appropriate scenarios. Also, we use the Mean-Decrease-in-Gini Parameter for laying these rules to obtain maximum homogeneity. The rules-based procedure is as follows:

```
DKA == GOOD
!       RS == ( GOOD) || (AVERAGE) : Accept
!       RS == POOR
!       !       PS == GOOD : Accept
!       !       PS == (AVERAGE) || (POOR) : Reject
DKA == AVERAGE
!       RS == GOOD : Accept
!       RS == AVERAGE
!       !       PS == GOOD
!       !       !       CS == GOOD : Accept
!       !       !       CS == (AVERAGE) || (POOR) : Reject
!       !       PS == (AVERAGE) || (POOR) : Reject
!       RS == POOR : Reject
DKA == POOR : Reject
```

Figure 3 Suggested procedure to be followed While Recruiting

As we see, during the selection process companies must initially clarify the values for each of the variable-parameters that are acceptable to them. For example some companies may accept students with average programming skills but others may only want those who have great programming skills whatever may be the scores for other attributes. Once that's done, they must start classifying students based on the features in the order of their variable importance values. Another important aspect here is to remember that in numerous scenarios, there may be a very large feature set. In such a case, it is not possible for the companies to consider all of them. Hence, to modify the selection tree, a tree pruning mechanism is developed as follows:

1. Set the threshold limit that must be used to prune to the tree, say 'P' percentage of pruning is needed.

2. Evaluate the percentage importance values using the following formula. If for variable $X_i$ the Mean-Decrease-in-Gini score is given as $Imp(X_i)$ then evaluate Percentage $Imp(X_i)$ as

$$Percent_{Imp(Xi)} = \frac{Imp(Xi)}{Imp(X1) + Imp(X2) + Imp(X3) + \ldots + Imp(Xn)} * 100$$

3. Now select the variables with minimum $Percent_{Imp(Xi)}$ scores and sum them until this total greater than or equal to 'P'.

$$Percent_{Imp(Xi)min} + Percent_{Imp(Xi+1)min-1} + Percent_{Imp(Xi+2)min-2} + \ldots Percent_{Imp(Xi+n)min-n} \geq P$$

4. All those variables $X_i$ with minimum $Percent_{Imp(Xi)}$ values that were added to the sum $\geq$ P should not be used during the construction of the ideal-tree.

This procedure allows us to cut down on the least important variables to be considered during the selection-procedure using the pre-pruning technique. Such a method would be helpful in scenarios when the number of variables is very high and it's not practical for the companies to look at every attribute. In case of a tie, it totally depends on the discretion of the hiring manager to exclude or include those features from the ideal-selection tree. Though this model provides high accuracy and allows pruning techniques to adapt to real-time scenarios, there is still scope for further improvement using cost optimization techniques like Gradient Boosting and many more, which would yield better results when dealing with higher dimensional data and inter-dependent variables. In order to develop a selection-framework for other domains, a similar procedure can be followed after incorporating minimal changes to suit them appropriately.

## VI. Conclusion

The primary target of data-mining is to produce near optimal results using the information extracted from patterns and trends hidden in historical data. This brings us to the essential question of choosing the best suitable model that can be applied to any given problem statement. In this scenario, the models of Decision trees have already been tested on a similar dataset with similar attributes and properties in [13]. However, this model suffers from the problem of overfitting and we try to overcome this issue by applying the Random-Forest technique. We also see that even when most studies provide a prioritization mechanism for selection procedure, they lack a quantifying measure to represent their importance in the framework.

The Random Forest method is a bagging based accuracy boosting technique, which creates multiple trees out of the bootstrap samples generated from a dataset, hence forming a forest. Here the output from each tree is considered to calculate an overall mean or an average result for the random forest. Since the calibration of the model is done using the out-of-bag samples, the model does not suffer from over-fitting issues which facilitates a superior performance in most of the scenarios when compared to Decision-trees. Apart from these enhancements in accuracy, Random forests also provide supplemental information like variable importance measures, etc. which adds to its value. One of these measures is the Mean Decrease in Gini index and Mean Decrease in Accuracy parameters which gives a holistic view about the contribution of each of the attributes to the final output. We utilize this facility provided by Random-Forest to assign an order of priorities to the features that must be considered during the recruitment of new-personnel.

We clearly see from the results that the GPA/grades of new recruits clearly aren't among the most important selection attributes on which a hiring-decision can be based. Other attributes like Programming Skills, Domain Specific Knowledge and Analytical Skills must be separately tested as they have a significant prediction power in estimating the performance of a person in the software industry. However, only obtaining a number of important parameters using a simple model do not suffice this need as the degree of accuracy associated with the model is equally important. This critical need for an accurate model would be clearly visible when developing a framework for other domains and working on high-dimensional data and this where Accuracy-boosting techniques like random forests come into play. These algorithms are not only known for their robustness but also perform well in cases of high-dimensional data and sparse data sets.Data-Mining algorithms extract important patterns and have helped in identification of project-members who have a greater probability to perform well. This research not only empowers the managers to refocus on Human-Capability criteria to enhance the development-process of any software-project but also address the issue where due to lack of analytical-methods in human-aspects, IT companies could not select the right talent in the software process and hence failed to achieve the desired-objectives in terms of both, quality and quantity in the time and cost-constraints. The scope of this study can be further extended to various domains and appropriate changes need to be incorporated when developing a selection-framework for each one of them. Such a study would require a preliminary research about finding the relevant features that can be directly correlated to the performance of any given employee in that particular domain. Once this is done, a similar model can be used to develop a feature-ranking order and enlist the most important attributes to be considered to make a hiring-decision using such a framework.

## Authors

**Gaurav Singh Thakur**
Gaurav Singh Thakur has completed his B.Tech in 2014 in Information Technology from National Institute of Technology Karnataka, Surathkal and is currently working as a Software Engineer at Cisco Systems, Inc. Bangalore. His technical areas of interest include Machine learning, Networking & Security, Application Development and Algorithms.

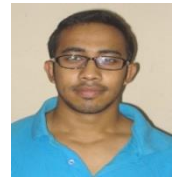

**Anubhav Gupta**
Anubhav Gupta has pursued his Bachelors, at National Institute of Technology Karnataka, Surathkal in the Field of Information Technology (IT) and graduated in 2014. His technical areas of interest include Machine learning, Information Security, Web Development and Algorithms. Currently, he is working as Software Developer Engineer at Commonfloor (MaxHeap Technologies).

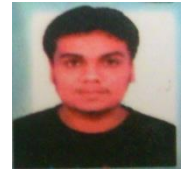

**Sangita Gupta**
Sangita Gupta is a research Scholar of Jain University Bangalore. She has done her B.Sc, M.Sc and M.Phil in computer science and guide for B.tech, M.Phil students

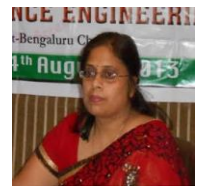